\documentclass[11pt]{article}

\usepackage[final]{acl}
\usepackage{times}
\usepackage{latexsym}
\usepackage[T1]{fontenc}
\usepackage[utf8]{inputenc}
\usepackage{microtype}
\usepackage{inconsolata}
\usepackage{graphicx}
\usepackage{booktabs}
\usepackage{amssymb}
\usepackage{enumitem}
\usepackage{amsmath}
\usepackage{placeins} 
\usepackage{multirow}
\usepackage{array}
\usepackage{fvextra}
\usepackage{dblfloatfix}  

\usepackage{bm}

\usepackage{hyperref}
\usepackage{xurl}

\title{Quantifying and Mitigating Socially Desirable Responding in LLMs:\\
A Desirability-Matched Graded Forced-Choice Psychometric Study}

\author{
  Kensuke Okada \\
  The University of Tokyo \\
  {\small \texttt{ken@p.u-tokyo.ac.jp}}
  \And
  Yui Furukawa \\
  The University of Tokyo \\
  {\small \texttt{yui-furukawa@p.u-tokyo.ac.jp}}
  \And
  Kyosuke Bunji \\
  Kobe University \\
  {\small \texttt{bunji@bear.kobe-u.ac.jp}}
}

\begin{document}
\maketitle

\begin{abstract}
Human self-report questionnaires are increasingly used in NLP to benchmark and audit large language models (LLMs), from persona consistency to safety and bias assessments. Yet these instruments presume honest responding; in evaluative contexts, LLMs can instead gravitate toward socially preferred answers---a form of socially desirable responding (SDR)---biasing questionnaire-derived scores and downstream conclusions. We propose a psychometric framework to quantify and mitigate SDR in questionnaire-based evaluation of LLMs. To quantify SDR, the same inventory is administered under HONEST versus FAKE-GOOD instructions, and SDR is computed as a direction-corrected standardized effect size from item response theory (IRT)-estimated latent scores. This enables comparisons across constructs and response formats, as well as against human instructed-faking benchmarks. For mitigation, we construct a graded forced-choice (GFC) Big Five inventory by selecting 30 cross-domain pairs from an item pool via constrained optimization to match desirability. Across nine instruction-following LLMs evaluated on synthetic personas with known target profiles, Likert-style questionnaires show consistently large SDR, whereas desirability-matched GFC substantially attenuates SDR while largely preserving the recovery of the intended persona profiles. These results highlight a model-dependent SDR–recovery trade-off and motivate SDR-aware reporting practices for questionnaire-based benchmarking and auditing of LLMs.\footnote{Data and code available at: \url{https://osf.io/2e6ny/}}
\end{abstract}

\section{Introduction}
\label{sec:intro}

Large language models (LLMs) are increasingly evaluated not only for task performance but also for higher-level behavioral and psychological profiles.
Benchmarks for cognitive capabilities and task performance in LLMs are now abundant. By contrast, despite the growing prominence of measuring and cultivating \emph{non-cognitive} traits in human development \citep{
HeckmanKautz2012HardEvidenceSoftSkills, OECD2021BeyondAcademicLearning}, methods to measure and evaluate analogous \emph{non-cognitive} attributes in LLMs---such as interpersonal style, value-laden tendencies, and stable response patterns---remain less standardized.

Psychometrics is the measurement science that develops and validates statistical models for inferring latent psychological traits from assessment responses. Since its early twentieth-century origins at the intersection of psychology and statistics \citep[e.g.,][]{Spearman1904GeneralIntelligence}, the field has established modern measurement frameworks, most notably item response theory 
\citep[IRT;][]{vanderLinden2018HIRTset}.

Recent LLM studies adapt psychometric inventories, such as Big Five personality questionnaires, as a pragmatic, theory-grounded tool for non-cognitive evaluation, assessing stable response-profile signatures in LLM outputs \citep{SerapioGarcia2023,Ye2025,Bhandari2025EvaluatingPersonalityLLMs}.
Throughout, we interpret questionnaire scores as \emph{behavioral} response tendencies under standardized prompts, rather than as evidence that LLMs possess human-like inner dispositions.
These studies open the door to \emph{AI psychometrics} \citep{Pellert2024,pellert2025neuralnetworkembeddingsrecover}, but they also inherit some challenges from human assessment.
A key challenge is \textit{socially desirable responding} (SDR), in which respondents tend to select options they believe are socially preferred or likely to impress evaluators, rather than those that reflect their true behavioral tendencies \citep{Paulhus2002}. This behavior can distort self-report results and reduce their validity, as inflated scores obscure genuine individual characteristics.

Recent studies have indeed observed that LLMs also produce 
questionnaire responses with a strong prosocial or idealized bias, much like those of a human respondent 
\citep{Bodroza2024, salecha2024social}. This is unsurprising, given that most LLMs are fine-tuned to be helpful and avoid offending, which is a form of \textit{alignment} with human preferences. 
Yet current LLM studies often report questionnaire-derived score profiles without quantifying the magnitude of SDR, without calibrating it to human benchmarks, and without deploying measurement formats designed to mitigate SDR.
This raises the question of whether—and to what extent—questionnaire-derived proxies of personality constructs in LLM outputs are biased by social desirability, and how to design evaluations that yield more robust, better-isolated measurements.

A substantial body of literature in human psychometrics shows that forced-choice formats reduce faking compared to single-stimulus Likert questionnaires, especially in high-stakes contexts \citep{
Cao2019,Martinez2021,Speer2023}. 
Crucially, this advantage requires IRT-based scoring of comparative data to avoid ipsativity and to place respondents on a common latent scale needed for our SDR quantification and mitigation. Under classical scoring, each forced-choice block allocates a fixed total of points across its options, producing constant-sum (ipsative) scores that confound absolute standing with within-profile trade-offs and therefore do not support valid between-respondent comparisons; IRT instead models comparative choices directly and recovers normative latent scores on a common scale \citep[see Section~\ref{sec:scoring_models} for more details; see also ][]{Brown2013}.
This paper transfers these principles to psychometric profiling of LLM outputs under questionnaire-style prompts.
Specifically, we treat SDR as a \emph{measurable response distortion} that can be quantified through instruction-induced contrasts and mitigated through desirability-controlled comparative measurement.
While design theory for desirability-matched comparative questionnaires is well developed, GFC Big Five instruments with explicit desirability matching are not commonly available.
To advance psychometrically grounded evaluation and mitigation of socially desirable responding in questionnaire-based LLM assessments, our contributions are threefold.

\paragraph{Contributions.}
\begin{enumerate}[leftmargin=*, topsep=0pt, itemsep=0pt, parsep=0pt, partopsep=0pt]

\item \textbf{Quantifying socially desirable responding:} Inspired by SDR quantification in human assessment, we propose a psychometrically grounded SDR metric for LLMs—an instruction-induced effect size computed on IRT-based latent scores.

\item \textbf{A desirability-matched GFC Big Five inventory:} We build a GFC Big Five inventory by selecting desirability-matched cross-domain statement pairs via constrained optimization grounded in forced-choice psychometrics.
\item \textbf{Empirical validation and exploratory model differences:} Across nine instruction-following LLMs, we show that desirability-matched GFC attenuates SDR relative to Likert while retaining ground-truth recovery (vs. persona targets), and we characterize how the remaining SDR--recovery trade-off varies across models.
\end{enumerate}

\section{Related Work}
\label{sec:related}
\paragraph{Questionnaire-based elicitation of behavioral response tendencies in LLMs.}
A growing body of work treats LLMs as respondents to psychometric questionnaires (e.g., Big Five) to characterize behavior and probe prompt-level controllability. Early studies profiled foundation models and compared scores to human norms \citep{miotto-etal-2022-gpt}. Subsequent work shows that profiles can be \emph{elicited} or \emph{steered} via prompting and can influence downstream text generation \citep{jiang-etal-2024-personallm}, but contextual cues alone can shift perceived personality, highlighting instruction- and context-dependence \citep{caron-srivastava-2023-manipulating}. A complementary line of work examines measurement \emph{reliability}: minor prompt or formatting perturbations can change responses substantially \citep{shu-etal-2024-dont,gupta-etal-2024-self}, though some stability is reported under extensive resampling \citep{huang-etal-2024-reliability}. Recent efforts propose LLM-oriented designs, including open-ended linguistic assessment \citep{zheng-etal-2025-lmlpa} and psychometrically validated multi-choice benchmarks \citep{lee-etal-2025-llms}.

\paragraph{Socially desirable responding in LLM survey settings.}
Response distortion in LLM questionnaires is closely related to \emph{sycophancy}-tuned assistants mirroring a user's stated beliefs rather than answering independently. \citet{perez-etal-2023-discovering} identify sycophancy via model-written evaluations, and \citet{sharma-etal-2024-sycophancy} analyze its prevalence and likely drivers. Although usually studied in open-ended dialogue, the same ``agree-with-the-user'' incentive is conceptually adjacent to survey settings: when a model infers what response is expected or preferred, it may favor socially palatable options over its baseline tendencies. In impression-management terms \citep{Paulhus2002}, sycophancy can be read as user-contingent desirability seeking, whereas SDR is its questionnaire-style analogue at the level of broad social norms. Direct evidence of socially desirable responding (SDR) in LLM assessment supports this. Across Big Five inventories, LLMs detect evaluation contexts and shift toward socially desirable profiles, paralleling human SDR \citep{salecha2024social}; broader AI psychometrics likewise finds interpretable yet systematically skewed profiles shaped by training data, safety tuning, and instruction framing \citep{Pellert2024}. 
However, existing SDR analyses mostly report aggregate score shifts, 
making it difficult to disentangle changes in inferred profile scores from social desirability and other response-style effects at the item-parameter level, which hinders calibrated comparisons across models, instruments, and human benchmarks.

\paragraph{Reducing response bias with comparative formats.}
A promising mitigation is to replace independent Likert ratings with \emph{comparative} judgments.
\citet{li-etal-2025-decoding-llm} compare Likert and forced-choice Big Five questionnaires for LLMs and find forced-choice to be less sensitive to temperature and to better differentiate models; they further explore self-reflection prompting and LLM-as-a-judge scoring.
Yet forced-choice formats can still admit desirability-driven selections unless the compared statements are carefully matched in desirability, and existing implementations often rely on heuristic scoring 
without psychometrically grounded model-based parameter estimation.

Taken together, prior work leaves open (i) how to quantify SDR in a way that is comparable across measured dimensions and anchored to human faking behavior, and (ii) how to design and score forced-choice inventories that reduce SDR while retaining interpretable profile scores.
Our work addresses these gaps with an IRT-calibrated SDR metric and a desirability-matched graded forced-choice inventory.

\section{Method}
\label{sec:method}

\subsection{Overview}
\label{sec:overview_notation}

\begin{figure*}[t]
  \centering
  \includegraphics[width=0.9\textwidth]{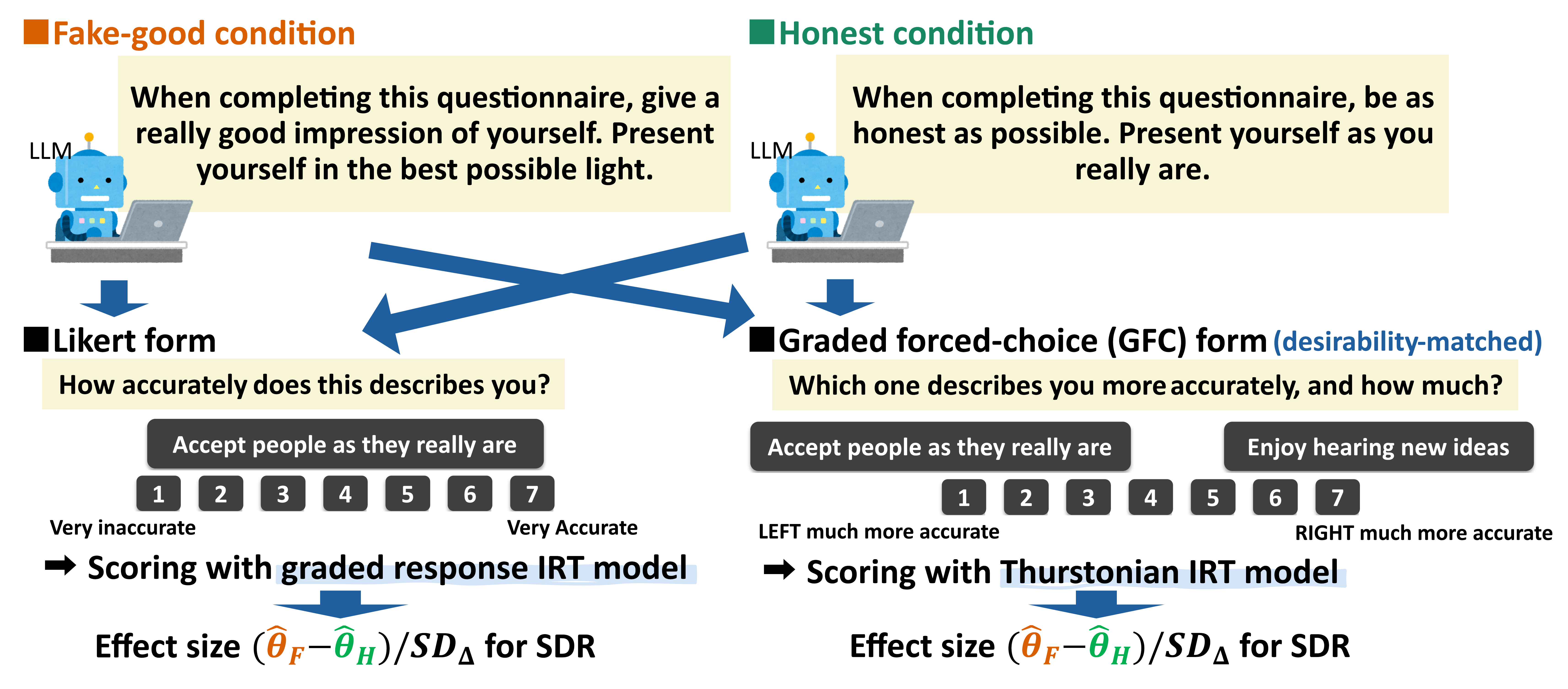}
\caption{Proposed framework to quantify and compare the socially desirable responding (SDR) of LLMs.}
  \label{fig:proposal}
\end{figure*}

Our method quantifies SDR as an instruction-induced shift in IRT-estimated scores. Concretely, we administer the same inventory under two instruction conditions: \textsc{HONEST} (answer as you really are) and \textsc{FAKE-GOOD} (answer to make the best possible impression). We fit an IRT model to place all responses on a common latent scale and define SDR as the resulting effect size (standardized difference) in Big Five profile scores between the two conditions (Figure \ref{fig:proposal}). This SDR effect size becomes the shared target quantity for our analyses: we test whether GFC responding suppresses desirability-driven distortion relative to standard Likert responding, and we conduct exploratory comparisons of SDR magnitudes across model families. 

To mitigate SDR, we first construct a desirability-matched Big Five GFC scale by estimating item desirability and pairing  statements from different Big Five domains  with closely matched desirability. This prevents models from uniformly selecting the ``socially best'' option and allows IRT scoring to recover interpretable latent profile scores under both HONEST and FAKE-GOOD conditions.

\subsection{Item pool, desirability estimation, and inventory construction}
\label{sec:item_pool}
\label{sec:inventory}
We begin from Goldberg's public-domain IPIP Big-Five factor-marker inventory \citep{Goldberg1999}, which contains 100 statements covering five domains.
Following our data-collection protocol, we exclude two voting-related statements and retain $J=98$ items.
Each statement $j$ is annotated with a Big Five domain label $f(j)\in\{A,C,E,N,O\}$ and a keying direction $g_j\in\{-1,+1\}$, where $A$, $C$, $E$, $N$, and $O$ denote Agreeableness, Conscientiousness, Extraversion, Neuroticism, and Openness to Experience, respectively.
$g_j=+1$ indicates that the statement is positively keyed (endorsing the statement indicates a higher latent score level), and $g_j=-1$ means it is negatively keyed.

A key ingredient for desirability matching is an item-level social desirability score $s_j$.
We estimate $s_j$ by prompting two strong LLMs (GPT-5 and Gemini~2.5~Pro) to act as ``desirability raters'' and judge how socially desirable the characteristic described by each statement is.
In this process, we use the same normative instruction framing and response anchors as prior human social desirability rating studies, asking how desirable each trait/characteristic is for an adult person on a 9-point scale \citep{Sankis1999GenderBias,CokerSamuelWidiger2002Maladaptive}. This gives $s_j$ a broad, norm-referenced interpretation for inventory construction---not as an attempt to approximate an ``LLM consensus,'' but as a desirability scale  grounded in a standard definition and elicitation protocol rather than an idiosyncratic provider-specific preference signal. We further assess the resulting ratings for within- and between-rater stability and for alignment with established human desirability norms \citep{britz2022elot}; the corresponding validation results are reported in Appendices~\ref{app:desirability} and~\ref{app:sd_validity}. These steps provide the desirability inputs needed for constructing the matched GFC inventory, illustrated schematically in Figure~\ref{fig:proposal} and specified in full in Appendix~\ref{app:gfc}.

\paragraph{Constructing desirability-matched graded forced-choice pairs.}
We construct a graded forced-choice inventory with $P=30$ paired-statement blocks (60 unique statements) by selecting \emph{cross-domain} item pairs whose desirability scores $s_j$ are as closely matched as possible.
Pair selection is posed as a constrained two-stage mixed-integer optimization that (i) minimizes the worst within-pair desirability gap and then (ii) among minimax-optimal solutions, minimizes overall mismatch while balancing domain coverage, domain-pair counts, and keying composition \citep{Sun2024FCConstruction,Li2025MixKeying}.
Appendix~\ref{app:gfc} provides the full formulation and the resulting inventory; the final set achieves a maximum within-block desirability gap of 0.18 on the 1--9 scale (mean 0.03).

\paragraph{Response formats.}
We administer the same 60 selected statements in two formats:
\begin{itemize}[leftmargin=*, topsep=0pt, itemsep=2pt, parsep=0pt, partopsep=0pt]
\item \textsc{Likert (single-stimulus):} each statement is rated on a 7-point Likert scale (1 = very inaccurate, 7 = very accurate).
\item \textsc{GFC (pairwise graded comparison):} statements are presented in 30 desirability-matched pairs; respondents provide a graded comparative judgment on a 7-point bipolar scale \citep[1 = LEFT much more accurate, 7 = RIGHT much more accurate; ][]{zhang_etal_2024_graded_forced_choice}.
\end{itemize}

\subsection{Psychometric scoring models}
\label{sec:scoring_models}
\paragraph{Why IRT scoring?}
To measure the response profiles of LLMs based on raw responses,
we use the IRT framework. 
The IRT offers a measurement backbone that is both theoretically grounded and practically useful, which is why it is adopted in major educational and psychological measurement programs such as OECD PISA \citep{OECD2024PISA2022TR}, NIH PROMIS \citep{PROMIS2013Standards,RoseEtAl2014PROMISPF}, and the SAT \citep{CollegeBoard2024DigitalSATFramework}.

IRT provides a principled \emph{measurement model} that maps observed responses to IRT-estimated latent scores via interpretable item parameters, rather than treating raw observed scores as direct measurements of the construct of interest.
It offers a broad set of benefits for measurement---from principled test scoring to uncertainty-aware latent profile estimation and scale linking \citep{vanderLinden2018HIRT3}. In our setting, one of the most important advantages is pragmatic and methodological: IRT provides a single, coherent measurement backbone that we can apply consistently across two different response formats, Likert ratings and GFC. This common backbone is essential for our goal of comparing socially desirable responding (SDR) across instruction conditions and LLM families without conflating true response distortion with format-specific scoring artifacts.

Notably, forced-choice responses cannot be treated as ordinary observed scores. Naively scoring comparative choices, such as  counting how often a statement or a choice option is selected within each block, produces \emph{ipsative}, constant-sum scores, which do not support valid between-respondent comparisons and can induce spurious dependencies among profile dimensions \citep{Brown2013}. Psychometrics addresses this limitation by extending IRT models from Likert-type data to forced-choice designs, enabling the recovery of normative latent scores from comparative judgments. In particular, Thurstonian IRT provides a direct solution to ipsativity by positing latent utilities for each statement and modeling observed choices as differences between these utilities \citep{BrownMaydeu2011FCIRT,Brown2016FCFramework,Buerkner2022ComparativeJudgments}. This yields latent score estimates on a normative scale that are comparable across respondents and conditions---exactly what is required to quantify SDR as a systematic shift in inferred profile scores rather than a byproduct of how responses are elicited.

\paragraph{Likert: five-dimensional graded response model (GRM).}
We score both formats with appropriate multidimensional IRT models fitted in Stan \citep{CarpenterEtAl2017Stan}.
Let $i$ index a response unit, which corresponds to a persona in our experiment 
(see Section~\ref{sec:exp_design}).
Let $\bm{\theta}_i \in \mathbb{R}^5$ denote its latent Big Five score vector.
Let $\bm{q}_j\in\{0,1\}^5$ be the one-hot vector indicating the domain measured by item $j$. 
Let $Y_{ij}\in\{1,\dots,7\}$ be unit $i$'s Likert response to item $j$.
We use a multidimensional graded response model \citep{Samejima1969GRM}, which is a standard IRT model for observed ordinal (Likert-type) item responses.
Define a signed discrimination $a_j = g_j a_j^{+}$ with $a_j^{+}>0$ and keying sign $g_j$.
The linear predictor is
\begin{equation}
\eta_{ij} = a_j \bm{q}_j^\top \bm{\theta}_i . \label{eq:grm1}
\end{equation}
With item-specific ordered thresholds $\bm{\kappa}_{j}=(\kappa_{j1},\dots,\kappa_{j6})$ satisfying
$
\kappa_{j1} < \kappa_{j2} < \cdots < \kappa_{j6},$ the GRM can be written as
\begin{align}
\Pr(Y_{ij} \ge k \mid \bm{\theta}_i) = \mathrm{logit}^{-1}\!\left(\eta_{ij}-\kappa_{j,k-1}\right),\nonumber \\
\quad k=2,\dots,7. \label{eq:grm2}
\end{align}

\paragraph{GFC: logistic ordinal Thurstonian IRT.}
For each pair $p\in\{1,\dots,30\}$ with the left statement $L_p$ and the right statement $R_p$, let $Y_{ip}\in\{1,\dots,7\}$ denote the graded comparative response.
Following the literature in the ordinal Thurstonian IRT model \citep{Brown2016FCFramework,BrownMaydeuOlivares2018OrdinalFA,Buerkner2022ComparativeJudgments,OkadaBunji2021Behaviormetrika}, we define each statement's latent utility
\begin{equation}
\mu_{ij} = a_j \bm{q}_j^\top \bm{\theta}_i, \label{eq:gfc1}
\end{equation}
and the comparative signal as the scaled right-minus-left difference
\begin{equation}
\eta_{ip} = \frac{\mu_{i,R_p}-\mu_{i,L_p}}{\sqrt{2}} .  \label{eq:gfc2}
\end{equation}
We then model the graded comparison with pair-specific ordered thresholds $\bm{\kappa}_p=(\kappa_{p1},\dots,\kappa_{p6})$ satisfying $
\kappa_{p1} < \kappa_{p2} < \cdots < \kappa_{p6}$:
\begin{align}
\Pr(Y_{ip} \ge k \mid \bm{\theta}_i) = \mathrm{logit}^{-1}\!\left(\eta_{ip}-\kappa_{p,k-1}\right),\nonumber \\
\quad k=2,\dots,7. \label{eq:gfc3}
\end{align}
so that larger $\eta_{ip}$ implies stronger endorsement of the RIGHT statement.

\subsection{Quantifying socially desirable responding}
\label{sec:sdr_metric}
Our SDR metric operationalizes instructed faking as an instruction-induced shift in latent score estimates.
For each LLM\footnote{For readability, we suppress the LLM index $l$ in what follows; all quantities are understood to be computed for a fixed LLM unless $l$ is shown explicitly.}, we obtain latent score estimates under \textsc{honest} and \textsc{fake-good} instructions for each response format $f\in\{\textsc{Likert},\textsc{GFC}\}$.
For each Big Five dimension $t \in \{A,C,E,N,O\}$, let $\hat{\theta}_{i,t,\textsc{honest}}$ and $\hat{\theta}_{i, t, \textsc{fake}}$ denote the posterior-mean latent score estimates. 

We define the raw within-persona shift for dimension $t$ as
\begin{equation}
\Delta_{i,t} = \hat{\theta}_{i,t,\textsc{fake}} - \hat{\theta}_{i,t,\textsc{honest}} .
\end{equation}
We compute the paired standardized effect size across personas (Cohen's $d_z$ for dependent means) \citep{Cohen1988Power,Lakens2013EffectSizes}:
\begin{equation}
d_{z,t} = \frac{\overline{\Delta}_{\cdot,t}}{SD_{\Delta_{i,t}}},
\end{equation}
where $\overline{\Delta}_{\cdot,t}$ is the mean shift across personas and $SD_{\Delta_{i,t}}$ is its standard deviation.

In Big Five measurement, higher Agreeableness, Conscientiousness, Extraversion, and Openness are generally socially desirable, whereas lower Neuroticism is desirable.
To make interpretation uniform across Big Five dimensions, we direction-correct signs so that positive values always indicate movement toward social desirability.
We define a desirability-direction multiplier $g_t$ as $g_t=+1$ for $t\in\{A,C,E,O\}$ and $g_t=-1$ for $t=N$.
Thus, our direction-corrected effect size is $\tilde d_{z,t}=g_t\, d_{z,t}$.

\subsection{Ground-truth recovery metric}
\label{sec:validity}
Because LLM ``personas'' have known ground-truth profiles by construction, we evaluate ground-truth recovery by correlating estimated profiles with persona specifications.
For each persona $i$, we have a target Big Five vector $\bm{z}_i=(z_{A,i},z_{C,i},z_{E,i},z_{N,i},z_{O,i})$ used to generate persona descriptions (Section~\ref{sec:persona_design}).
For response format $f$ and instruction condition $c\in\{\textsc{honest},\textsc{fake-good}\}$, we compute the Pearson correlation between the estimated profile vector $\hat{\bm{\theta}}_{i,c}$ and the target $\bm{z}_i$ across personas, and report dimension-wise and averaged correlations.
This captures whether measurement preserves rank-order differences in intended personality profiles.

\section{Experimental Setup}
\label{sec:experiments}

\subsection{Personas and ground-truth profiles}
\label{sec:persona_design}
To mimic an LLM adopting diverse user personas and to evaluate how well the IRT models recovered the intended profiles, 
we generated a shared set of 50 synthetic personas with known Big Five target values and used it for all LLMs.
We sample $\bm{z}_i$
from a multivariate standard normal distribution with mean zero and an empirical human Big Five correlation structure adopted from \citet{vanderLinden2010GFP} so that the sampled personas resemble realistic joint distributions of real-world Big Five profiles (Appendix~\ref{app:persona}).
We then map $\bm{z}_i$ to trait adjectives by selecting descriptors from a curated trait lexicon, ensuring coverage across the five domains.

For constructing natural-language descriptions that are more amenable to processing by LLMs, we transformed each profile score into persona descriptions based on the 
52 markers from \citet{SerapioGarcia2023}. Specifically, for each domain, we produced one sentence listing adjectives/phrases corresponding to the assigned stanine level, using intensity terms such as ``extremely,'' ``very,'' or ``a bit'' depending on that domain's stanine score.  Each persona description always covered all five dimensions and was framed by an explicit role instruction (``YOU ARE THE RESPONDENT'') and a final directive to answer as that person would.
In this way, we achieved both the assignment of ground-truth Big Five scores and the imitation of adopted personas that are realistic and amenable to LLMs.

\subsection{Experimental design, models, and prompting}
\label{sec:exp_design}
We evaluate nine instruction-following LLMs from three providers, accessed via their official APIs: OpenAI (GPT-5, GPT-5 mini, GPT-5 nano), Google (Gemini 2.5 Pro, Gemini 2.5 Flash, Gemini 2.5 Flash-Lite), and Anthropic (Claude Opus 4.5, Claude Sonnet 4.5, Claude Haiku 4.5); exact model identifiers/snapshots are listed in Appendix~\ref{app:model_specs}.
For each persona and LLM, we administer both response formats under a fully crossed within-respondent design with two factors:
(i) instruction condition (\textsc{honest} vs.\ \textsc{fake-good}) and
(ii) response format (\textsc{Likert} vs.\ \textsc{GFC}).
Thus each persona yields four questionnaire response sets per model.

\paragraph{Instruction manipulation and prompting.}
To elicit SDR, we administer each questionnaire under two instruction conditions.
In the \textsc{honest} condition, the prompt instructed the persona to answer as honestly as possible. 
In the \textsc{fake-good} condition, the prompt instructed the persona to answer in order to give a really good impression and to present themselves in the best possible light. These instructions were adopted from \citet{Furnham1997KnowingFaking}; the resulting \textsc{honest}--\textsc{fake-good} contrast follows a standard instructed-faking design in human SDR research, used to elicit and quantify intentional response distortion \citep{Paulhus2002,Speer2023}. Because the design varies only the instructional framing while holding the persona description and questionnaire constant, this contrast is expected, by design, to preferentially isolate desirability-driven variance. In this sense, the \textsc{fake-good} manipulation serves as a controlled stress test of SDR and supports comparison with human instructed-faking benchmarks \citep{Speer2023}.
Although we use \textsc{fake-good} as a controlled stress test, the same SDR metric and IRT scoring can be applied to more naturalistic questionnaire settings in which no explicit faking prompt is given but alignment-tuned models still treat the survey as evaluative and drift toward socially desirable profiles.

Prompts present the persona description, followed by the questionnaire items or pairs, and require the model to output only the requested numeric responses.
Full prompts are provided in Appendix~\ref{app:prompts}.

\subsection{Data collection and IRT estimation}
\label{sec:data_collection}
We collect complete response vectors for each (LLM, persona, format, condition) combination.
Within each persona, item/pair order is randomized once and held fixed across instruction conditions to support within-persona comparisons.
All LLMs are queried via their official APIs in batch mode using each provider's default generation/decoding configuration (see Appendix~\ref{app:model_specs}). 
We fit the GRM and ordinal Thurstonian IRT models in Stan with weakly informative priors and standard identifiability constraints (Appendix~\ref{app:stan}).

\begin{figure*}[t]
  \centering
  \includegraphics[width=\textwidth]{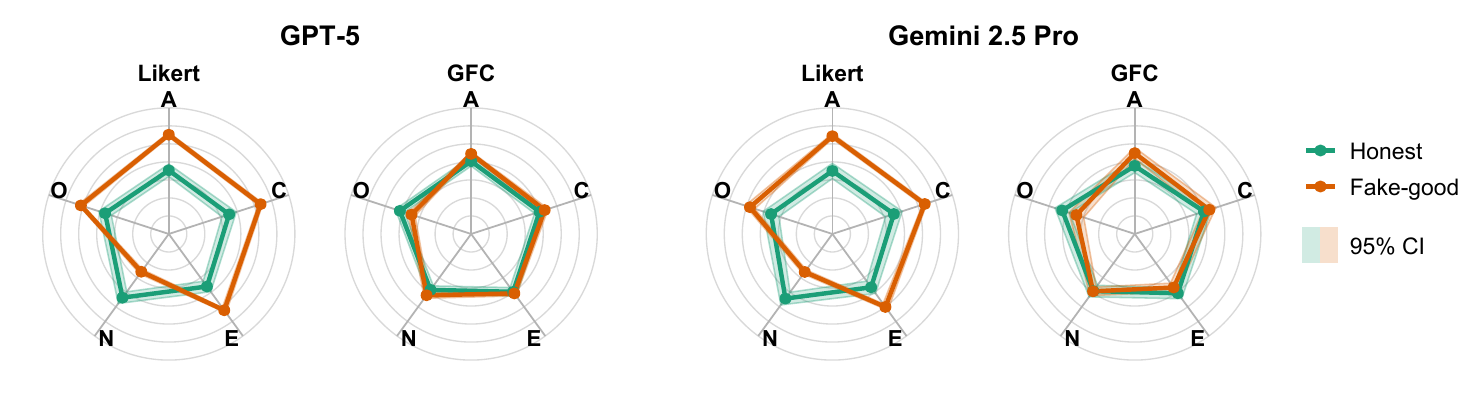}
\caption{Mean ($\pm 95\%$ CI) Big Five profiles for two representative high-capacity LLMs (GPT-5 and Gemini 2.5 Pro)
under \textsc{honest} vs.\ \textsc{fake-good} instructions. For each model, profile estimates are shown separately for Likert responses
(left) and graded forced-choice (GFC) responses (right).}
  \label{fig:radar_profiles}
\end{figure*}


\section{Results}
\label{sec:results}

\subsection{Profile distortion under fake-good instructions}

Figure~\ref{fig:radar_profiles} visualizes the fake-good distortion observed in our experiments for GPT-5 and Gemini~2.5~Pro, showing how the response format shapes the inferred Big Five profile.
Under the single-stimulus Likert format, fake-good instructions shift the inferred Big Five profile toward a stereotypically ``good person'' pattern: higher A, C, E, and O, and lower N. The fake-good (orange) polygon expands on socially valued Big Five dimensions and contracts on Neuroticism relative to the honest (green) polygon.
By contrast, the GFC panels show substantially smaller separations between honest and fake-good profiles for the same models. When an LLM must make graded comparative judgments between (approximately) equally desirable statements, it becomes difficult to uniformly endorse all ``good'' content. As a result, the honest and fake-good profiles in GFC are much closer than their Likert counterparts.

\begin{figure*}[t]
  \includegraphics[width=\textwidth]{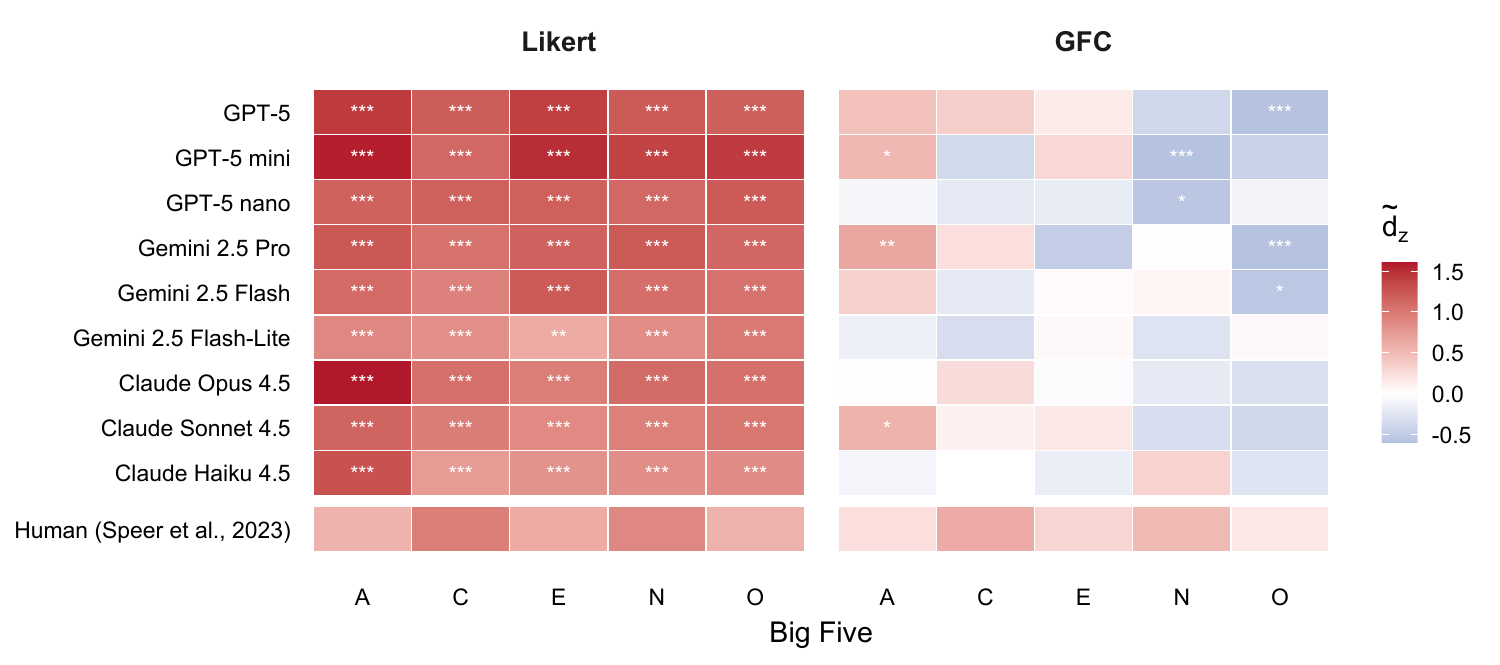}
\caption{SDR effects on Big Five profile estimates (direction-corrected Cohen's $\tilde{d}_z$ computed on latent $\hat{\theta}$).
Positive values indicate shifts toward the socially desirable direction (higher A, C, E, and O; lower N). 
Asterisks indicate Bonferroni-corrected significance across all cell-wise HONEST--FAKE-GOOD comparisons (* $p<.05$, ** $p<.01$, *** $p<.001$).
The Likert panel shows consistently large, positive SDR across models and dimensions, whereas the GFC panel shows substantially attenuated (often near-zero) effects. The human row (Speer et al., 2023) is shown as a descriptive benchmark only and is not significance-tested.}
  \label{fig:theta_effects}
\end{figure*}

\begin{figure*}[t]
  \centering
  \includegraphics[width=\textwidth]{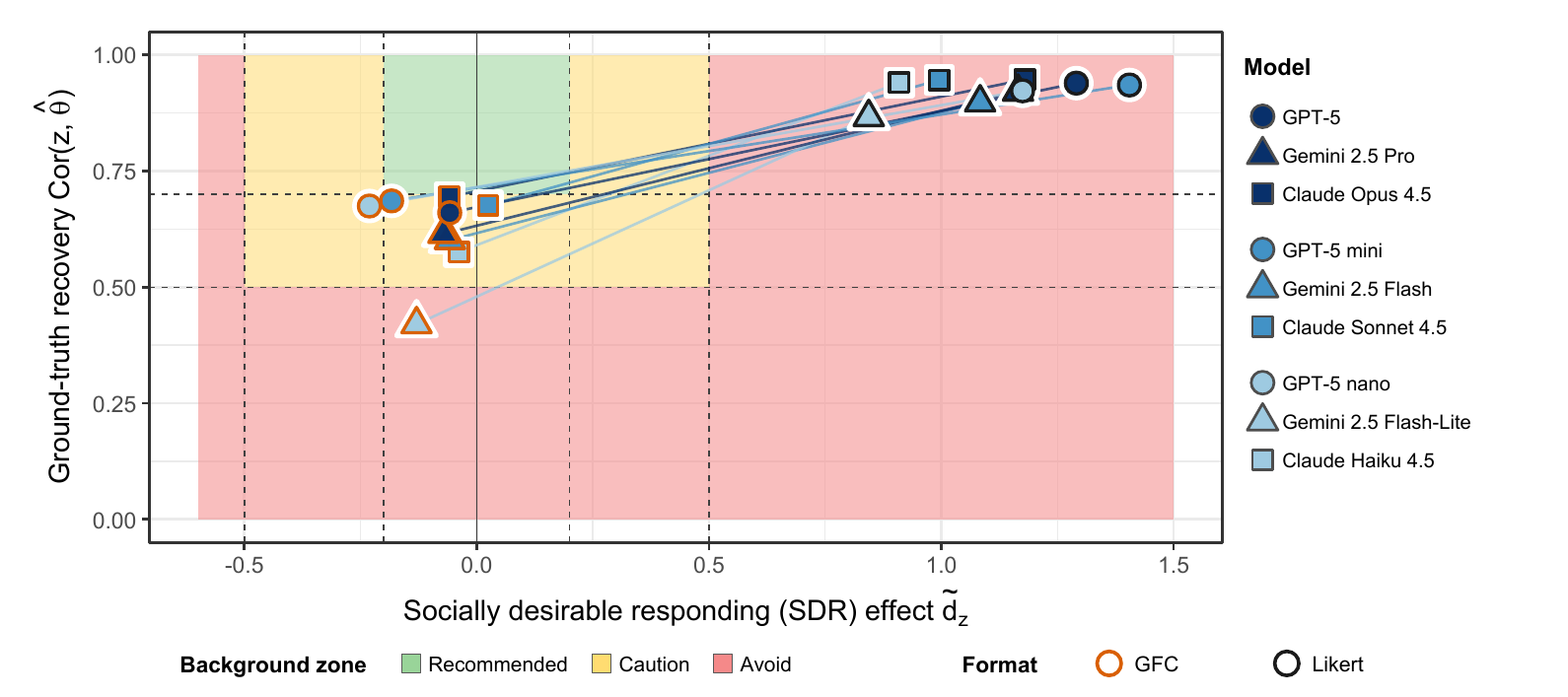}
\caption{SDR--recovery trade-off across response formats.
For each LLM and format, we plot summary SDR shift (direction-corrected Cohen's $\tilde{d}_z$) against ground-truth recovery (Pearson $r$ between true persona scores $z$ and IRT-estimated $\hat{\theta}$) under honest responding, both aggregated across dimensions.
Marker shape encodes model family, blue fill shade encodes the provider-designated capacity variant within each family, and border color encodes response format (GFC vs.\ Likert).
Thin connector lines link the two response formats within a model. Background colors indicate practical interpretation zones: for SDR, $|\tilde{d}_z|\le 0.2$ (\citeauthor{Cohen1988Power}'s (\citeyear{Cohen1988Power}) ``small'' effect) is considered a practically negligible shift (recommended), $0.2<|\tilde{d}_z|\le 0.5$ (up to ``medium'') as caution but potentially acceptable, and $|\tilde{d}_z|>0.5$ as avoid; these cutoffs are adopted from the equivalence-testing literature discussing equivalence bounds (smallest effect size of interest, SESOI) for assessing practically negligible differences \citep{LakensScheelIsager2018Equivalence}.
For recovery, following the convergent-validity literature, we interpret $r\ge 0.70$ as strong (recommended), $0.50\le r<0.70$ as acceptable/moderate, and $r<0.50$ as insufficient \citep{AbmaEtal2016Convergent}.}
\label{fig:sdr_validity_tradeoff}
\end{figure*}

\subsection{SDR magnitude under Likert vs.\ desirability-matched GFC}
\label{sec:sdr_results}

To quantify the magnitude of SDR across models, Figure~\ref{fig:theta_effects} summarizes the effect sizes on the latent Big Five score estimates ($\hat{\theta}$) for each response format. The heatmap reports direction-corrected Cohen's $d_z$ effect sizes \citep[fake-good minus honest; ][]{Cohen1988Power,Lakens2013EffectSizes}, which are paired-samples standardized mean differences oriented so that positive values always indicate a shift toward the socially desirable direction (higher A, C, E, O, and lower N). 
Because our goal is to quantify SDR magnitude on a shared comparative metric, we treat direction-corrected Cohen's $\tilde{d}_z$ as the primary quantity; asterisks in Figure~\ref{fig:theta_effects} provide supplementary Bonferroni-corrected significance from cell-wise paired-samples $t$-tests on the within-persona HONEST--FAKE-GOOD contrasts. Consistent with the effect-size pattern, all 45 Likert LLM cells are significant at the 1\% level, whereas under GFC only 4/45 remain significant.

The Likert panel of Figure~\ref{fig:theta_effects} shows a strikingly consistent pattern: for every evaluated model, fake-good instructions produce positive shifts on all five dimensions once aligned to desirability. 
This indicates that Likert responses allow LLMs to substantially reshape their inferred personality profiles when prompted to ``look good.'' In other words, on Likert items the fake-good manipulation behaves like a global response-style shift toward desirability rather than a subtle or domain-specific perturbation.
Descriptively, these magnitudes are comparable to human instructed-faking effects summarized in meta-analyses \citep{Speer2023}.

In contrast, desirability-matched GFC substantially attenuates SDR across most models.
The reduction is strongest for models that show the largest Likert SDR, suggesting that desirability matching and comparative judgment jointly constrain impression management.
However, some models retain non-negligible SDR even under GFC, indicating that comparative formats do not fully eliminate desirability-driven response distortion in LLMs.

\subsection{Recovery of persona ground truth}
\label{sec:validity_results}
SDR attenuation alone is not sufficient because a response format could trivially yield near-zero SDR by suppressing domain-relevant variance, leaving the resulting profile estimates uninterpretable.
Because our personas have ground-truth target profile vectors by construction, we therefore check whether estimated scores preserve this signal via correlations with the persona ground truth.

Figure~\ref{fig:sdr_validity_tradeoff} summarizes the resulting SDR--recovery relationship.
Under Likert, models often achieve higher ground-truth recovery, but their SDR shifts are consistently large, placing most points in the red (avoid) region---meaning that in self-presentational contexts, the inferred profiles can be substantially distorted by desirability-driven responding.
The dominant effect of our proposed desirability-matched GFC is therefore the large SDR reduction: for every model, switching from Likert to GFC produces a pronounced leftward shift, typically moving into the yellow/green (negligible-to-moderate SDR) region.
This mitigation comes with a cost in ground-truth recovery, but the drop is generally limited: most GFC points remain in the acceptable-to-strong convergent-validity bands ($r\ge 0.50$), indicating that GFC suppresses desirability-driven distortion while largely preserving profile signal.

We therefore use the randomly assigned persona target profiles as the central baseline for evaluation, rather than an unprompted ``default'' condition. Because each synthetic persona comes with an explicit ground-truth Big Five target, this baseline lets us ask not only whether a format attenuates SDR, but also whether it preserves the intended behavioral signal. Without such an explicit target, a response format could appear to reduce SDR simply by collapsing responses toward the mean. The practical implication is therefore not ``always use GFC'' but \emph{SDR-aware evaluation}: Likert may be preferable when maximizing recovery of a prompted persona under strictly honest conditions, whereas desirability-matched GFC is preferable in evaluative settings---such as comparative auditing or safety/fairness/value surveys---where self-presentational pressure can distort Likert profiles.

Although the general pattern is consistent, the degree of trade-off also varies by model family.
These patterns imply that alignment and response policies interact with measurement format in model-specific ways.
We qualitatively examine whether SDR magnitude and SDR--recovery trade-offs track coarse provider-designated capacity variants within providers (e.g., GPT-5 / mini / nano). While higher-capacity variants sometimes show lower SDR under GFC, the relationship is not uniform: some smaller models show minimal additional SDR beyond baseline, while some larger models retain measurable SDR under both formats.
These findings motivate future work on linking SDR to specific training objectives and safety policies.

\section{Conclusion}
\label{sec:conclusion}

We proposed a psychometric framework to quantify and mitigate SDR in questionnaire-based profiling of LLM outputs. Building on SDR frameworks used in human psychometrics, we quantify SDR as a direction-corrected instruction-induced effect size on IRT-based profile scores. To reduce SDR, we develop a desirability-matched graded forced-choice Big Five inventory scored with ordinal Thurstonian IRT. Across nine instruction-following LLMs, Likert responses exhibit large SDR, whereas desirability-matched GFC substantially reduces SDR while largely preserving ground-truth recovery. Together, these findings motivate SDR-aware reporting practices for questionnaire-based LLM evaluation in computational linguistics: whenever questionnaire-style prompts are used for comparative auditing or for safety, fairness, and value evaluation, studies benefit from reporting both inferred profiles and their susceptibility to social-desirability distortion.

\FloatBarrier
\raggedbottom
\section{Limitations}
\label{sec:limitations}

\paragraph{Empirical contrasts and format comparison.}
In Figure~\ref{fig:theta_effects}, the human reference row is taken from a meta-analysis \citep{Speer2023} that aggregates heterogeneous instruments, scoring approaches, and study contexts; as a result, the reported human effects do not necessarily reflect a single, desirability-matched GFC design scored with the same Thurstonian IRT model used in our study, and desirability is not always explicitly controlled in the primary studies.
By contrast, our LLM results come from a single, tightly controlled design using the same statement pool, a desirability-matched GFC construction, and a shared IRT metric.
Consequently, qualitative human--LLM differences should be interpreted cautiously: they may reflect methodological heterogeneity in addition to intrinsic human--LLM differences.
In this respect, our LLM experiments arguably provide a more controlled benchmark, but a direct, design-matched human--LLM study is needed to make stronger comparative claims.

In addition, we matched the \emph{number of statements} across formats, but this implies that Likert collects twice as many responses as GFC (60 ratings vs.\ 30 pair judgments), potentially disadvantaging GFC in our ground-truth recovery comparison.
Future work should therefore (i) administer the same desirability-matched inventory to humans and LLMs and score both with the same Thurstonian IRT model, and (ii) compare formats under matched response counts or matched test information.

\paragraph{Scope and practical relevance.}
Our claims are primarily scoped to questionnaire-style benchmarking under context-aware persona-steering scenarios.
Here, we use \emph{persona steering} broadly to include both direct prompt conditioning (e.g., system/user prompts or explicit profile injection) and indirect conditioning conveyed through product- or UI-level scaffolding once it becomes part of the model's effective context.
This is narrower than LLM evaluation in general, but it still covers deployment-relevant systems in which models are guided toward particular roles or behavioral profiles, including AI assistants, tutors, mental-health chatbots, and gaming NPCs.
In such settings, implicit socially desirable responding from safety tuning may distort or partially override intended system profiles and thereby obscure model differences in behavioral auditing.

\paragraph{Instrument and ground truth.}
Our study focuses on a specific personality instrument \citep[IPIP Big-Five factor-marker inventory;][]{Goldberg1999}.
Although this inventory and the Big Five profiles measured by it are considered representative of non-cognitive traits, other inventories may yield different SDR patterns.
Our personas provide a controlled ground truth for recovery evaluation, but they are synthetic and may not reflect the full richness of human trait expression. More implicit or narrative persona specifications (e.g., dialogue histories) may diffuse profile signals and make the SDR--recovery trade-off more pronounced. 
Finally, we treat each persona--instruction instance as a separate response unit when fitting IRT; future work could explicitly model within-persona dependence and condition effects in a hierarchical framework.

\section{Ethical Considerations}
This work studies SDR in questionnaire-style evaluation of LLMs and introduces a desirability-matched GFC inventory to reduce impression-management effects. We intend the method as an auditing and measurement tool for questionnaire-based LLM evaluation, including safety, fairness, and value surveys where evaluative prompting can mask model differences.

The main ethical risk is dual use: the same design principles could be used not only to detect SDR but also to support evaluation evasion or less detectable self-presentation. We therefore frame the method as a research and auditing tool and caution against using questionnaire-derived profiles for high-stakes decisions about people. In addition, desirability judgments are culturally and context dependent and may encode rater-specific or majoritarian norms. Applications to new languages, cultures, or domains should therefore re-estimate desirability ratings and consider stakeholder-specific norms. Our study uses synthetic personas and does not involve human participants or personal data; any extension to human assessment should require separate validation, consent, and appropriate ethics oversight.

\section*{Acknowledgments}
This work was supported by JST AIP Acceleration Research Grant Number JPMJCR25U2 and JSPS KAKENHI Grant Number 25H00577.
We used AI assistance (ChatGPT 5.2--5.4) to polish
the writing of the manuscript, as well
as to refine the code for our analyses and visualizations.

\bibliography{llm_faking}

\appendix

\section{Estimating social desirability scores for IPIP items}
\label{app:desirability}

\paragraph{Overview.}
We estimate item-level social desirability scores $s_j$ (where $j \in \{1,\dots,100\}$ ) for the IPIP item pool by prompting two rater LLMs (GPT-5 and Gemini~2.5~Pro) to rate how socially desirable each statement is for an adult person.
The instruction framing and the 1--9 desirability anchors follow standard human social desirability rating protocols used in lexical and evaluative-norm studies (\citealt{Sankis1999GenderBias};  \citealt{CokerSamuelWidiger2002Maladaptive}).

\paragraph{Prompt and Response Format.}
Each rating query presented the model with a block of 25 personality statements.
We adopt the same normative instruction wording and desirability anchors as in human ratings \citep{Sankis1999GenderBias,CokerSamuelWidiger2002Maladaptive}, and add only the strict ``return integers only'' constraint to support automatic parsing.
The general instruction (shared across blocks and rater models) read:

\begin{quote}
\small
The following statements are characteristics of people. Indicate on a scale from 1 to 9 \textbf{how desirable} you think each trait or characteristic is for an adult person: 1 = Very undesirable, 3 = Undesirable, 5 = Neutral, 7 = Desirable, 9 = Very desirable. Use any number from 1 through 9 (i.e., 1, 2, 3, 4, 5, 6, 7, 8, or 9) that best indicates your opinion regarding how desirable that trait is.

Please return EXACTLY 25 integers separated by single spaces, in the SAME ORDER as the statements.

Do not include any other text.
\end{quote}

Below this instruction, the 25 items in the current block were listed line by
line in the form \texttt{Statement: \textit{[sentence]}} and visually delimited
from the instruction text by marker lines (``\texttt{++++}''). Thus, the desired
output for each query was a single line consisting of 25 integers in
$\{1,\dots,9\}$, each representing the judged desirability of the corresponding
statement.

\paragraph{Randomization and Batching.}
For each rater model
$l \in \{\text{GPT-5}, \text{Gemini 2.5 Pro}\}$ and each replication
$r \in \{1,\dots,30\}$, we performed the following steps:

\begin{enumerate}[leftmargin=*, topsep=2pt, itemsep=0pt, parsep=0pt, partopsep=0pt]
\item We drew a random permutation of the $J$ statements without replacement.
  \item The permuted sequence was partitioned into four consecutive blocks of
        25 items each (blocks $b=1,\dots,4$).
  \item For each block $b$, we constructed a prompt as described above and
        submitted it to the rater model via the provider API using batch
        processing (OpenAI via \texttt{ellmer::batch\_chat\_text}; Gemini via the
        Gemini Batch API over REST).
\end{enumerate}

Each prompt was issued as a single-turn request (no chat history). Across 30
replications and 4 blocks per replication, each rater model therefore evaluated
$30 \times 4 = 120$ prompts and responded $120 \times 25 = 3000$ desirability ratings.

\paragraph{Quality Control and Refitting of Non-conforming Outputs.}
Raw API responses were parsed and quality-checked to ensure they could be
unambiguously aligned to the 25 statements in the corresponding prompt. In our
implementation, we normalized each response by removing line breaks, commas,
and whitespace, and then verified that the resulting string consisted of
exactly 25 single-digit ratings and contained only digits in $\{1,\dots,9\}$.
Responses failing this check were flagged as non-conforming.

For flagged prompts, we re-issued the query once using the same item block but
with an augmented prefix reminding the model that (i) this is a psychometric
rating task, (ii) some items may mention sensitive topics, and (iii) it must
return only the requested integers in the requested format. At this step, there were no responses that failed the format check.

\paragraph{Aggregation to Item-level Desirability Scores.}
Because the order of statements was randomized on every replication, we stored
the ordered list of item identifiers included in each 25-item prompt and aligned
the returned ratings back to item IDs by position (1--25). Let $x_{jlr}$ denote
the rating assigned to item $j \in \{1,\dots,100\}$ by rater model $l$ on
replication $r$. We computed the final desirability score for item $j$ as the
mean of all available ratings,
\[
  s_j = \frac{1}{60} \sum_{l} \sum_{r} x_{jlr},
\]
We used the mean $s_j$ as the
desirability input to the optimization procedure described in Section~\ref{sec:inventory}.

\paragraph{Agreement of observed desirability ratings.}
To check the consistency between ratings, treating items as targets and replications as raters, we computed a two-way random-effects, absolute-agreement intraclass correlation coefficient (ICC; ICC(A,1) for a single replication and ICC(A,30) for the mean across 30 replications).
Within each model, agreement was extremely high (Gemini 2.5 Pro: ICC(A,1)=$.980$; GPT-5: ICC(A,1)=$.975$), and reliability of the 30-replication mean exceeded $.999$ for both models (Table~\ref{tab:desirability-consistency}).
Complementary indices supported the same conclusion (mean pairwise replication correlations $\approx .98$; split-half reliability $>.998$).
Moreover, agreement between LLMs on per-item mean ratings was also very high (Pearson $r=.993$, ICC(A,1)=.989), indicating that observed desirability ratings were stable both within and across models.

\begin{table*}[t]
\centering
\small

\begin{tabular}{lcccc}
\hline
\multicolumn{5}{l}{\textbf{Within-LLM consistency across 30 independent replications (100 items)}} \\
\hline
LLM & ICC(A,1) & ICC(A,30) & Mean pairwise $r$ &  Split-half $r$ \\
\hline
Gemini 2.5 Pro & .980 [.974, .985] & .9993 [.9991, .9995] & .980 &  .9986 [.9981, .9991] \\
GPT-5          & .975 [.968, .982] & .9992 [.9989, .9994] & .976 &  .9984 [.9978, .9988] \\
\hline
\end{tabular}

\vspace{0.6em}

\begin{tabular}{lccc}
\hline
\multicolumn{4}{l}{\textbf{Between-LLM agreement on per-item mean ratings (100 items)}} \\
\hline
Comparison & Pearson $r$ & Spearman $\rho$ & ICC(A,1) \\
\hline
Gemini 2.5 Pro vs.\ GPT-5 & .993 [.989, .995] & .987 & .989 [.984, .993]\\
\hline
\end{tabular}

\caption{Agreement of observed desirability ratings.
ICC(A,1) and ICC(A,30) denote two-way absolute-agreement ICCs for a single replication and the 30-replication mean, respectively (brackets: 95\% confidence intervals).
Mean pairwise $r$ is the mean off-diagonal correlation among replications across items.
Split-half $r$ is based on 2,000 random 15/15 splits of replications (brackets: 2.5/97.5 percentiles).
}
\label{tab:desirability-consistency}
\end{table*}

\section{Convergent Validity of LLM-based Social Desirability Ratings}
\label{app:sd_validity}

\paragraph{Motivation.}
Our scale construction relies on social desirability estimates $s_j$ obtained from LLM raters
(Appendix~\ref{app:desirability}) to form desirability-matched forced-choice blocks.
A potential threat to validity is that LLM-based desirability judgments might deviate
substantially from human norms, in which case desirability matching could be miscalibrated.
We therefore conducted an external convergent-validity check by comparing LLM-based ratings
to established human social desirability norms for trait words.

\paragraph{Human benchmark.}
We used the social desirability norms from \citet{britz2022elot}, which provide mean social
desirability ratings (\texttt{SOC\_MEAN}) for 500 English trait adjectives collected from large
human samples.
Let $w \in \{1,\dots,100\}$ index a random sample of $W=100$ adjectives drawn without replacement
from this database, and let $y_w \in [-3,3]$ denote the corresponding human mean desirability
rating (with $-3$ = very undesirable and $+3$ = very desirable).

\paragraph{LLM ratings (10 independent repetitions).}
For each sampled adjective $w$, we obtained ratings from the same two rater LLMs used in
Appendix~\ref{app:desirability} (GPT-5 and Gemini~2.5~Pro).
Each repetition queried the LLM in a single-turn request (no chat history) with the same
normative instruction framing as in Appendix~\ref{app:desirability} (rating ``how desirable'' a
trait is for an adult person) and the 9-point integer response scale $\{1,\dots,9\}$.
Appendix A indicates that the LLM ratings were highly consistent. Therefore, we collected $R=10$ repetitions per LLM, each time shuffling the 100 words and batching them into
four prompts of 25 words, yielding up to $100\times 10 = 1000$ raw ratings per LLM.

\paragraph{From raw outputs to per-word means.}
For each LLM $l$ and repetition $r$, let $x_{wlr}\in\{1,\dots,9\}$ denote the parsed desirability
rating returned for word $w$ (missing if the response could not be aligned to the prompt after a
single refit attempt).
To avoid inflating the effective sample size, we first aggregate repetitions at the word level.
Specifically, for each word $w$ and LLM $l$, we compute the per-word mean $\bar{x}_{wl}$
so that each word contributes a single LLM value to the correlation analysis.

\paragraph{Correlation computation.}
For each LLM $l$, Table~\ref{tab:llm_human_soc_corr} reports (i) the Pearson product--moment
correlation between the vectors $\{\bar{x}_{wl}\}_{w=1}^{W}$ and $\{y_w\}_{w=1}^{W}$ and (ii) the
Spearman rank correlation between the same vectors.
As shown in this table, the LLM-based desirability ratings closely track
the human social desirability norms.

\begin{table}[t]
\centering
\small
\begin{tabular}{lcc}
\toprule
Rater LLM & Pearson $r$ (95\% CI) & Spearman $\rho$ \\
\midrule
GPT-5            & 0.950 \,[0.927, 0.966] & 0.943 \\
Gemini~2.5~Pro   & 0.950 \,[0.926, 0.966] & 0.947 \\
\bottomrule
\end{tabular}
\caption{Correlations between per-word mean LLM desirability ratings $\bar{x}_{wl}$ 
and human mean social desirability ratings $y_w$ from \citet{britz2022elot}, computed across the sampled
$W=100$ trait adjectives.}
\label{tab:llm_human_soc_corr}
\end{table}

\section{Construction of the Graded Forced-Choice Inventory}
\label{app:gfc}

\paragraph{Item pool and desirability inputs.}
We begin from Goldberg's public-domain IPIP Big-Five factor-marker inventory \citep{Goldberg1999}, which contains 100 statements covering five domains.
At this stage, we decided to exclude two voting-related statements from the original item pool and retained $J=98$ statements. This is because voting behavior is highly susceptible to construct-irrelevant influences—political attitudes/ideology, cross-country institutional differences, and eligibility differences, thereby confounding Big Five measurement.
Each statement $j$ is annotated with a Big Five domain label $f(j)\in\{A,C,E,N,O\}$ and a keying sign $g_j\in\{+1,-1\}$, where $g_j=+1$ denotes a positively keyed item and $g_j=-1$ a negatively keyed item.
The social desirability score $s_j$ for each item is the LLM-based estimate described in
Appendix~\ref{app:desirability}.

\paragraph{Candidate cross-domain pairs.}
We considered only cross-domain pairs, i.e., pairs of statements that measure different Big Five dimensions.
Let $\mathcal{P}=\{(j,j'): 1\le j<j'\le J,\ f(j)\neq f(j')\}$ be the set of all such unordered pairs, and let
$x_{jj'}\in\{0,1\}$ indicate whether pair $(j,j')\in\mathcal{P}$ is included in the final inventory.
For each candidate pair $(j,j')$, define the absolute desirability gap $\Delta_{jj'}=|s_j-s_{j'}|$ and a mixed-key indicator
$g_{jj'}=\mathbb{I}[g_j\neq g_{j'}]$.

\begin{table*}[!tbp]
\centering
\scriptsize
\setlength{\tabcolsep}{3pt}
\renewcommand{\arraystretch}{1.05}
\begin{tabular}{r c p{0.31\textwidth} r c p{0.31\textwidth} r r}
\toprule
BLK & Dom. (key) & Statement & SD & Dom. (key) & Statement & SD & $|\Delta SD|$ \\
\midrule
1 & A+ & Accept people as they are. & 8.75 & O+ & Enjoy hearing new ideas. & 8.75 & 0.00 \\
2 & A+ & Am concerned about others. & 8.87 & C+ & Carry out my plans. & 8.87 & 0.00 \\
3 & A+ & Am easy to satisfy. & 6.63 & E+ & Talk to a lot of different people at parties. & 6.65 & 0.02 \\
4 & A+ & Believe that others have good intentions. & 7.43 & O+ & Can say things beautifully. & 7.62 & 0.18 \\
5 & A- & Contradict others. & 3.23 & O- & Am not interested in abstract ideas. & 3.17 & 0.07 \\
6 & A- & Cut others to pieces. & 1.00 & N+ & Dislike myself. & 1.02 & 0.02 \\
7 & A- & Get back at others. & 1.03 & C- & Shirk my duties. & 1.03 & 0.00 \\
8 & A+ & Have a good word for everyone. & 8.08 & N- & Am not easily bothered by things. & 8.07 & 0.02 \\
9 & A- & Have a sharp tongue. & 2.58 & E- & Retreat from others. & 2.62 & 0.03 \\
10 & A+ & Respect others. & 9.00 & N- & Remain calm under pressure. & 9.00 & 0.00 \\
11 & A- & Suspect hidden motives in others. & 2.57 & E- & Would describe my experiences as somewhat dull. & 2.53 & 0.03 \\
12 & A+ & Treat all people equally. & 9.00 & C+ & Complete tasks successfully. & 9.00 & 0.00 \\
13 & C+ & Am exacting in my work. & 7.92 & O+ & Have a rich vocabulary. & 7.80 & 0.12 \\
14 & C+ & Do things according to a plan. & 7.93 & O+ & Have a vivid imagination. & 7.93 & 0.00 \\
15 & C- & Don't put my mind on the task at hand. & 1.65 & N+ & Often feel blue. & 1.63 & 0.02 \\
16 & C- & Don't see things through. & 1.52 & N+ & Am often down in the dumps. & 1.50 & 0.02 \\
17 & C- & Find it difficult to get down to work. & 1.95 & N+ & Get stressed out easily. & 1.93 & 0.02 \\
18 & C+ & Get chores done right away. & 8.07 & E+ & Make friends easily. & 8.07 & 0.00 \\
19 & C+ & Make plans and stick to them. & 8.75 & E+ & Cheer people up. & 8.68 & 0.07 \\
20 & C- & Need a push to get started. & 2.78 & E- & Find it difficult to approach others. & 2.83 & 0.05 \\
21 & C+ & Pay attention to details. & 8.17 & O+ & Enjoy thinking about things. & 8.10 & 0.07 \\
22 & E+ & Am skilled in handling social situations. & 8.57 & N- & Rarely lose my composure. & 8.57 & 0.00 \\
23 & E+ & Am the life of the party. & 6.47 & O+ & Enjoy wild flights of fantasy. & 6.45 & 0.02 \\
24 & E- & Avoid contacts with others. & 2.17 & N+ & Fear for the worst. & 2.17 & 0.00 \\
25 & E- & Don't talk a lot. & 4.98 & O- & Believe that too much tax money goes to support artists. & 4.90 & 0.08 \\
26 & E- & Keep in the background. & 4.70 & O- & Do not like poetry. & 4.63 & 0.07 \\
27 & E- & Keep others at a distance. & 2.70 & N+ & Am filled with doubts about things. & 2.72 & 0.02 \\
28 & N+ & Worry about things. & 2.70 & O- & Have difficulty understanding abstract ideas. & 2.63 & 0.07 \\
29 & N- & Rarely get irritated. & 8.32 & O+ & Get excited by new ideas. & 8.33 & 0.02 \\
30 & N- & Seldom get mad. & 8.05 & O+ & Carry the conversation to a higher level. & 8.07 & 0.02 \\
\bottomrule
\end{tabular}
\caption{The final 30-block graded forced-choice inventory constructed by two-stage desirability matching.
Big Five domain labels: A=Agreeableness, C=Conscientiousness, E=Extraversion, N=Neuroticism, O=Openness to experience.
``+''/``--'' indicate item keying.
SD is the LLM-estimated social desirability score on a 1--9 scale; $|\Delta SD|$ is the absolute within-block difference.}
\label{tab:gfc-inventory}
\end{table*}


\paragraph{Two-stage (lexicographic) mixed-integer optimization.}
We selected exactly $P=30$ pairs.
To obtain uniformly tight desirability matching, we solved a two-stage mixed-integer program.
In the first stage we minimized the maximum within-pair desirability gap by introducing a continuous variable $m\ge 0$:
\[
  \min_{x,m}\ m
\]
subject to
\begin{align}
  \sum_{(j,j')\in\mathcal{P}} x_{jj'} &= P, \label{eq:gfc-npairs}\\
  \Delta_{jj'} x_{jj'} &\le m \qquad \forall (j,j')\in\mathcal{P}, \label{eq:gfc-minimax}\\
  \sum_{\substack{(j,j')\in\mathcal{P}:\\ j=\ell\ \text{or}\ j'=\ell}} x_{jj'} &\le 1 \qquad \forall \ell\in\{1,\dots,J\}, \label{eq:gfc-unique}\\
  0.4P &\le \sum_{(j,j')\in\mathcal{P}} g_{jj'}x_{jj'} \le 0.6P, \label{eq:gfc-mixedkey}
\end{align}
and the following balance constraints.
First, we enforced equal representation of the five domains by requiring each Big Five domain to appear
exactly $2P/5=12$ times:
\begin{equation}
\begin{aligned}
  &\sum_{(j,j')\in\mathcal{P}}
  \mathbb{I}[f(j)=t\ \text{or}\ f(j')=t]\ x_{jj'}
  = 12\\
  &\qquad \forall t\in\{A,C,E,N,O\}.
\end{aligned}
\label{eq:gfc-domain}
\end{equation}
Second, we balanced the \emph{domain-pair} composition by requiring each of the ten unordered domain pairs
to occur exactly $P/10=3$ times:
\begin{equation}
\begin{aligned}
  &\sum_{(j,j')\in\mathcal{P}}
  \mathbb{I}[\{f(j),f(j')\}=\{t,t'\}]\ x_{jj'}
  = 3\\
  &\qquad \forall \{t,t'\}\subset\{A,C,E,N,O\},\ t<t'.
\end{aligned}
\label{eq:gfc-domainpair}
\end{equation}
Finally, to avoid extreme imbalances in positive/negative keying within any domain, let
\begin{align}
  N_{t,+} &=
  \sum_{(j,j')\in\mathcal{P}}
  \Bigl(
  \mathbb{I}[f(j)=t, g_j=+1] \notag\\
  &\qquad + \mathbb{I}[f(j')=t, g_{j'}=+1]
  \Bigr) x_{jj'},\\
  N_{t,-} &=
  \sum_{(j,j')\in\mathcal{P}}
  \Bigl(
  \mathbb{I}[f(j)=t, g_j=-1] \notag\\
  &\qquad + \mathbb{I}[f(j')=t, g_{j'}=-1]
  \Bigr) x_{jj'},
\end{align}
and enforced that each sign accounts for at least 30\% of the selected items within each domain:
\begin{equation}
\begin{aligned}
  7N_{t,+} &\ge 3N_{t,-}\ \ \text{and}\ \ 7N_{t,-}\ge 3N_{t,+}\\
  &\qquad \forall t\in\{A,C,E,N,O\}.
\end{aligned}
\label{eq:gfc-signbalance}
\end{equation}

Let $m^\star$ denote the optimal value of the first-stage problem.
In the second stage, we kept the minimax optimum by adding the constraint $m\le m^\star+\varepsilon$ (with $\varepsilon=10^{-9}$)
and minimized the total squared desirability mismatch:
\begin{align*}
  \min_{x,m}\quad & \sum_{(j,j')\in\mathcal{P}} (s_j-s_{j'})^2 x_{jj'}\\
  \text{s.t.}\quad & \text{Equations~(\ref{eq:gfc-npairs})--(\ref{eq:gfc-signbalance}) and } m\le m^\star+\varepsilon.
\end{align*}
We implemented this optimization in \texttt{R} using \texttt{ompr}/\texttt{ROI} and solved both stages with the Gurobi solver.

\paragraph{Resulting inventory.}
The final inventory contains 30 two-statement blocks (60 unique items).
The optimized solution attained a maximum within-block desirability gap of 0.18 (mean 0.03, SD 0.04; range [0.00, 0.18]) on the 1--9 scale.
Table~\ref{tab:gfc-inventory} lists the resulting pairs, together with their LLM-estimated desirability scores.

\section{Full IRT model specifications}
\label{app:IRTmodels}\label{app:stan}
This appendix documents the Bayesian prior specification, identifiability constraints, and Stan implementation details for the IRT models described in Section~\ref{sec:scoring_models}.
The likelihood components are given in Equations~(\ref{eq:grm1})--(\ref{eq:grm2}) (Likert GRM) and (\ref{eq:gfc1})--(\ref{eq:gfc3}) (GFC ordinal Thurstonian IRT).

\paragraph{Response units and pooling across LLMs/conditions.}
For each response format, we fit a single pooled IRT model across all LLMs and both instruction conditions (Section~\ref{sec:scoring_models}, ``A shared latent metric across LLMs and conditions'').
Each row $i\in\{1,\dots,N\}$ of the response matrix corresponds to one completed questionnaire under a fixed (LLM $l$, persona $r$, condition $c$) combination.
Before model fitting, we excluded incomplete response sets (i.e., rows with missing item/pair responses) to keep the likelihood well-defined.

\paragraph{$Q$-matrix and keying.}
The design matrix $Q\in\{0,1\}^{J\times D}$ is one-hot, so that each item loads on exactly one Big Five domain.
Reverse-keyed items are handled by a sign vector $g\in\{-1,+1\}^J$, which flips the direction of the item discrimination.

\paragraph{Parameterization.}
In both the Likert and GFC models we parameterize the signed item discrimination as
$a_j=g_j a_j^{+}$ with $a_j^{+}>0$.
In Stan this is implemented by sampling a positive ``strength'' parameter $a_j^{+}$ and multiplying by $g_j$ inside the likelihood.
Ordered category thresholds are represented by an \texttt{ordered[K-1]} vector, enforcing strict ordering
$\kappa_{j1}<\cdots<\kappa_{j,K-1}$ (Likert) and $\kappa_{p1}<\cdots<\kappa_{p,K-1}$ (GFC).

\paragraph{Priors and scale identification.}
We use weakly informative priors that are identical across the two response formats:
\begin{align}
  \bm{\theta}_i &\sim \mathcal{N}(0, \mathbf{I}
  _5), \qquad i=1,\dots,N, \\
  a_j^{+} &\sim \mathcal{N}^{+}(0,0.5), \qquad j=1,\dots,J, \\
  \kappa_{j,k} &\sim \mathcal{N}(0,1.5), \qquad j=1,\dots,J, \notag\\
  &\qquad k=1,\dots,K-1, \\
  \kappa_{p,k} &\sim \mathcal{N}(0,1.5), \qquad p=1,\dots,P, \notag\\
  &\qquad k=1,\dots,K-1.
\end{align}
Here $\mathcal{N}^{+}(0,0.5)$ denotes a half-normal prior (a Normal$(0,0.5)$ distribution truncated to $a_j^{+}>0$).
The multivariate standard normal prior $\bm{\theta}_i\sim\mathcal{N}(0,\bm{I}_5)$ fixes the latent location and scale (mean 0, variance 1 for each dimension) and assumes prior independence across the five dimensions; we do not estimate a latent correlation matrix in the IRT scoring step.

\paragraph{Link to Stan's \texttt{ordered\_logistic} parameterization.}
In Stan we implement both models with \texttt{ordered\_logistic($\eta$, $\kappa$)}.
Stan's ordered-logistic cutpoint convention is
$\Pr(Y\le k)=\mathrm{logit}^{-1}(\kappa_k-\eta)$.
This is algebraically equivalent to the GRM form in Equation~(\ref{eq:grm2}) because
$\Pr(Y\ge k)=1-\Pr(Y\le k-1)=\mathrm{logit}^{-1}(\eta-\kappa_{k-1})$.

\paragraph{GFC scaling convention.}
Following Equation~(\ref{eq:gfc2}), the GFC linear predictor uses the standardized utility difference
$\eta_{ip}=(\mu_{i,R_p}-\mu_{i,L_p})/\sqrt{2}$.
In Stan this is implemented by multiplying the right-minus-left difference by a constant $1/\sqrt{2}$ (stored as \texttt{inv\_sqrt2}).
This keeps the a priori scale of the comparison signal comparable to the single-item GRM predictor (a difference of two independent utilities would otherwise have roughly twice the prior variance).

\paragraph{Stan implementation and computation.}
We fit both models in Stan using \texttt{cmdstanr}.
To reduce computation, we precompute $\bm{Q}^\top$ once and form the matrix $\bm{Z}=\bm{\Theta} \bm{Q}^\top$ so that $Z_{ij}=\bm{q}_j^\top\bm{\theta}_i$; likelihood contributions are then evaluated item-wise (GRM) or pair-wise (GFC) with vectorization over response units.
Both Stan programs contain no \texttt{generated quantities}; all derived metrics (SDR indices, correlations, and visualizations) are computed in \texttt{R} from posterior summaries.

\paragraph{MCMC settings, diagnostics, and posterior summaries.}
We ran 4 chains, with 200 warmup iterations and 500 post-warmup iterations per chain, using NUTS with \texttt{adapt\_delta}=0.95 and \texttt{max\_treedepth}=12 and a fixed random seed.
Sampling was parallelized across chains via \texttt{cmdstanr}; to avoid oversubscription we set \texttt{OMP\_NUM\_THREADS=1} (no within-chain OpenMP parallelism).
We used Stan defaults for parameter initialization and did not apply thinning.
Convergence was monitored using $\widehat{R}$ (threshold 1.01) and effective sample sizes.
For downstream analyses we use posterior means of $\theta$ as point estimates, denoted $\hat{\theta}$ in the main text.

\section{Persona generation details}
\label{app:persona}

We draw $\bm{z}_i$ independently from a five-dimensional Gaussian distribution with
zero mean and covariance $\Sigma$:
\begin{equation}
\bm{z}_i \stackrel{\mathrm{i.i.d.}}{\sim} \mathcal{N}_5(\bm{\mu}, \Sigma),
\qquad
\bm{\mu}=\mathbf{0}\in\mathbb{R}^5.
\end{equation}
We set $\Sigma$ to the meta-analytic corrected Big Five intercorrelation matrix
reported by \citet[Table~2]{vanderLinden2010GFP}, reordered to match our domain
ordering $(A,C,E,N,O)$:
\begin{equation}
\Sigma
=
\begin{pmatrix}
1.00 & 0.43 & 0.26 & -0.36 & 0.21 \\
0.43 & 1.00 & 0.29 & -0.43 & 0.20 \\
0.26 & 0.29 & 1.00 & -0.36 & 0.43 \\
-0.36 & -0.43 & -0.36 & 1.00 & -0.17 \\
0.21 & 0.20 & 0.43 & -0.17 & 1.00
\end{pmatrix}.
\end{equation}

\section{Questionnaire administration prompts}
\label{app:prompts}

This appendix records the \emph{exact} prompt text used to elicit questionnaire responses, as implemented in our data-collection script.
Each API call used a single user message that concatenates a persona prefix (Appendix~\ref{app:persona}), an instruction block (\textsc{honest} or \textsc{fake-good}), and either Likert statements or GFC statement pairs.
Angle-bracket placeholders (e.g., \texttt{<STATEMENT>}) indicate run-time substitutions and are \emph{not} literal tokens sent to the models.

\subsection{Persona prefix (prepended to every request)}
\begin{small}
\begin{Verbatim}[breaklines=true,breaksymbolleft={},breaksymbolright={}]
YOU ARE THE RESPONDENT.

<TRAIT_SENTENCE_O>
<TRAIT_SENTENCE_C>
<TRAIT_SENTENCE_E>
<TRAIT_SENTENCE_A>
<TRAIT_SENTENCE_N>

Answer all questions AS THIS PERSON would.
\end{Verbatim}
\end{small}

\noindent where each \texttt{<TRAIT\_SENTENCE\_*>} is a single English sentence describing the persona (e.g., ``You are very \dots.''), derived from the sampled ground-truth trait vector (Appendix~\ref{app:persona}).

\subsection{Instruction blocks}
The instruction block was appended immediately after the persona prefix.

\paragraph{\textsc{honest} instruction.}
\begin{small}
\begin{Verbatim}[breaklines=true,breaksymbolleft={},breaksymbolright={}]
You will complete a personality questionnaire. When completing this questionnaire,
we would like you to be as honest as possible, that is, to present yourself as you really are.
\end{Verbatim}
\end{small}

\paragraph{\textsc{fake-good} instruction.}
\begin{small}
\begin{Verbatim}[breaklines=true,breaksymbolleft={},breaksymbolright={}]
You will complete a personality questionnaire. When completing this questionnaire,
we would like you to answer the questions in order to give a really good impression of yourself,
that is, to present yourself in the best possible light.
\end{Verbatim}
\end{small}

\subsection{Likert (7-point) prompt}

\begin{small}
\begin{Verbatim}[breaklines=true,breaksymbolleft={},breaksymbolright={}]
<PERSONA_PREFIX>

<INSTRUCTION_BLOCK>

Indicate how accurately the following statement describes you
using a 7-point scale:
1: Very Inaccurate
2: Moderately Inaccurate
3: Slightly Inaccurate
4: Neither Accurate nor Inaccurate
5: Slightly Accurate
6: Moderately Accurate
7: Very Accurate
Return ONLY one integer (1-7).
Do not include any other text.
++++
Statement: <STATEMENT>
++++
\end{Verbatim}
\end{small}

\subsection{GFC (7-point bipolar) prompt}

\begin{small}
\begin{Verbatim}[breaklines=true,breaksymbolleft={},breaksymbolright={}]
<PERSONA_PREFIX>

<INSTRUCTION_BLOCK>

For the following pair of statements, indicate which one describes you
more accurately and by how much
using a 7-point bipolar scale:
1: LEFT statement describes me much more accurately
2: LEFT statement describes me moderately more accurately
3: LEFT statement describes me slightly more accurately
4: About the same
5: RIGHT statement describes me slightly more accurately
6: RIGHT statement describes me moderately more accurately
7: RIGHT statement describes me much more accurately
Return ONLY one integer (1-7).
Do not include any other text.
++++
LEFT: <LEFT_STATEMENT>  ||  RIGHT: <RIGHT_STATEMENT>
++++
\end{Verbatim}
\end{small}

\paragraph{Left/right randomization.}
For each GFC pair, the assignment of the two statements to the \texttt{LEFT} vs.\ \texttt{RIGHT} slot was randomized; the numeric anchors above therefore always refer to the \emph{displayed} left/right positions.

\subsection{Output acceptance and retries}
A response was accepted if (i) it contained the expected number of integers for that request and (ii) all integers lay in $\{1,\dots,7\}$.
Requests failing these checks (including outputs with extra text that prevented extracting the expected count) were re-submitted with the identical prompt up to three additional times; at this point, there were no remaining failures.

\FloatBarrier
\section{Model identifiers, snapshots, query dates, and decoding settings}
\label{app:model_specs}

Table~\ref{tab:model_specs} reports the exact model identifiers/snapshots for the experiments
reported in this paper. 
\begin{table}[!t]
\centering
\small
\begin{tabular}{ll}
\toprule
Label in text & Model identifier / snapshot \\
\midrule
 GPT-5                  & \texttt{gpt-5-2025-08-07} \\
GPT-5 mini             & \texttt{gpt-5-mini-2025-08-07} \\
 GPT-5 nano             & \texttt{gpt-5-nano-2025-08-07} \\
Gemini 2.5 Pro         & \texttt{gemini-2.5-pro} \\
Gemini 2.5 Flash       & \texttt{gemini-2.5-flash} \\
Gemini 2.5 Flash-Lite  & \texttt{gemini-2.5-flash-lite} \\
Claude Opus 4.5        & \texttt{claude-opus-4-5-20251101} \\
Claude Sonnet 4.5      & \texttt{claude-sonnet-4-5-20250929} \\
Claude Haiku 4.5       & \texttt{claude-haiku-4-5-20251001} \\
\bottomrule
\end{tabular}
\caption{Model identifiers/snapshots for the experiments reported in this paper.}
\label{tab:model_specs}
\end{table}
GPT-5 and Gemini~2.5~Pro were additionally used as the two desirability raters in
Appendix~A; the remaining seven models were used only in the questionnaire collection stage.
Query dates were constant within stage: the desirability-rating stage was run on 2025-12-14--15,
and the questionnaire collection stage was run on 2025-12-24--25. Across all providers and both
stages, our scripts passed no explicit sampling or decoding overrides; all runs therefore used the
providers' default generation/decoding settings.

\end{document}